\documentclass{article}
\usepackage{spconf,amsmath,graphicx,booktabs}
\usepackage[hidelinks]{hyperref}
\usepackage{xcolor}

\newcommand{\EmotionVCRetention}{36.5}

\newcommand{\CountVCRetention}{2.1}

\newcommand{\MandarinCER}{7.53}
\newcommand{\MandarinRate}{7.7}

\newcommand{\MandarinCharacterRate}{3.2}

\newcommand{\MandarinPhoneRate}{7.1}

\title{ASSET: Adaptive Speech Segmentation via Reinforcement Learning}
\name{\begin{tabular}{@{}c@{}}
Cihan Xiao$^{1}$, Yuan Tseng$^{2}$, Matthew Wiesner$^{3}$, Desh Raj$^{4}$,\\
Leibny Paola Garcia Perera$^{1, 3}$, Sanjeev Khudanpur$^{1, 3}$, David Harwath$^{2}$
\end{tabular}}
\address{$^{1}$Johns Hopkins University, USA \quad $^{2}$The University of Texas at Austin, USA\\
$^{3}$HLTCOE, Johns Hopkins University, USA \quad $^{4}$NVIDIA, USA}
\hypersetup{pdfauthor={Cihan Xiao, Yuan Tseng, Matthew Wiesner, Desh Raj, Leibny Paola Garcia Perera, Sanjeev Khudanpur, David Harwath}}

\begin{document}
\ninept
\maketitle

\begin{abstract}

Speech large language models (LLMs) are often built using pretrained acoustic encoders as an audio interface to an LLM decoder. However, LLMs scale poorly with input sequence length, while acoustic encoders tend to produce substantially longer sequences than typical textual inputs. This motivates learning to compress encoder outputs while preserving task-oriented information. To this end, we introduce ASSET, a reinforcement learning (RL)-based framework that learns to adaptively compress encoder outputs by first partitioning them into variable-duration segments and then pooling each segment into a continuous audio token. We use group-relative policy optimization (GRPO) to train a segmenter with task performance feedback and a token-rate penalty. In automatic speech recognition (ASR) experiments, ASSET achieves performance comparable to a transcript-assisted character-boundary oracle with approximately 25\% fewer audio tokens on English LibriSpeech, while improving on a phone-boundary oracle at a similar token rate on Mandarin AISHELL. Exploratory experiments on non-ASR tasks further illustrate how learned segmentation changes when the framework is adapted to tasks beyond transcription.
\end{abstract}


\section{Introduction}
\label{sec:introduction}

Processing audio with LLMs is challenging, as such systems usually combine a pretrained LLM decoder~\cite{bai2024seed,xu2026firered,10447605,wu2023decoder,zhou2025cjst} with an acoustic encoder which typically produces high-frequency feature sequences that are expensive to process.
The resulting long audio prefixes increase context usage and the cost of decoder self-attention, which grows quadratically with input sequence length.
When the downstream task does not require the encoder's full temporal resolution, some of this computation may be unnecessary.
Our goal is therefore to learn how to compress encoder outputs into shorter sequences while preserving task-oriented acoustic information.

\begin{figure*}[t]
  \centering
  \includegraphics[width=\textwidth]{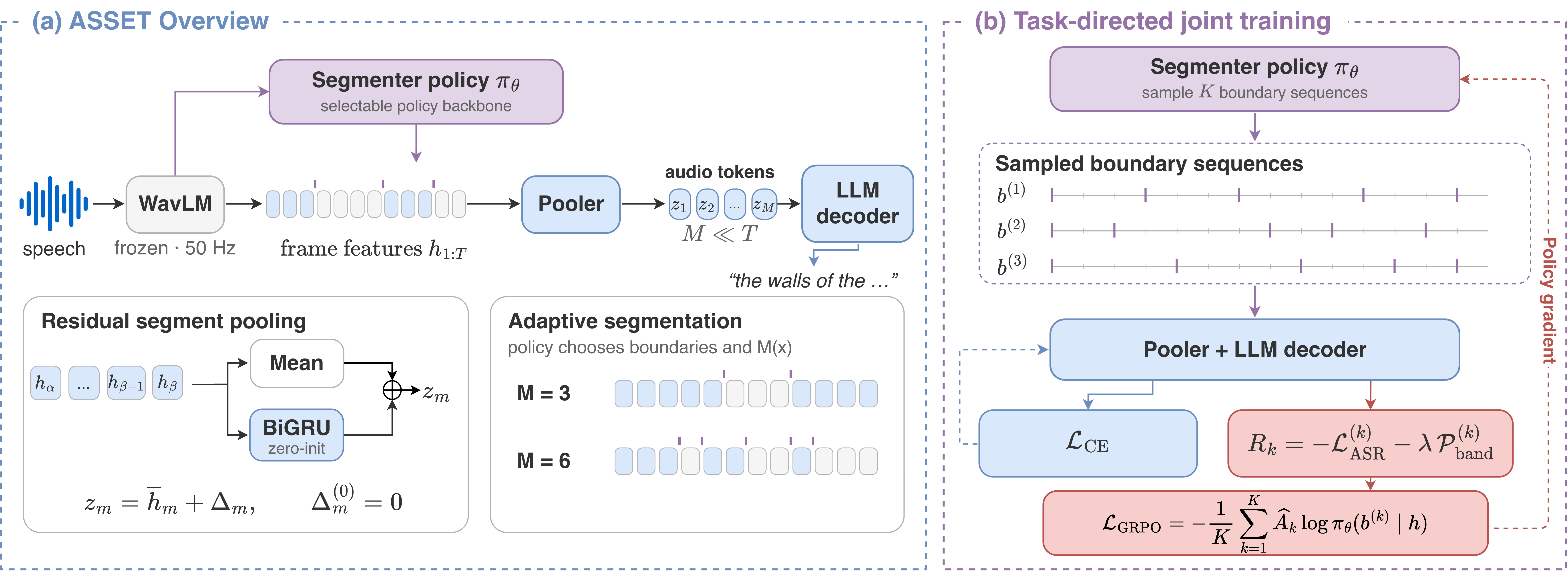}
  \caption{ASSET overview. \textbf{(a)} A policy partitions encoder frames into variable-duration segments, and a residual BiGRU pooler maps each segment to an audio token. \textbf{(b)} Task losses score sampled boundary sequences; cross-entropy updates the pooler and decoder, while GRPO updates the segmenter.}
  \label{fig:asset_overview}
\end{figure*}

We view frame compression as two complementary operations: \emph{segmentation}, which partitions the frame sequence into contiguous segments, and \emph{pooling}, which summarizes each segment into one audio token (Fig.~\ref{fig:asset_overview}). Fixed-rate frame stacking or pooling uses evenly distributed boundaries \cite{bhati2025compression}, whereas connectionist temporal classification (CTC) compression uses auxiliary alignments for variable-rate token allocation~\cite{zhou2025cjst,graves2006ctc}.
CTC provides a natural criterion for ASR, however, for tasks such as emotion recognition, a suitable auxiliary alignment objective is less obvious. This motivates learning compression directly from the target task.

Both segmentation and pooling can be learned, but they require different strategies. A pooling function can be optimized via backpropagation directly once segment assignments are given. However, boundary decisions are discrete: moving a boundary changes which frames belong to each segment, blocking ordinary backpropagation through these assignments. RL provides a convenient way to train such a boundary policy model by scoring how its segmentation policies affect the downstream task performance, without requiring an auxiliary frame-level target.
In this paper, we introduce \textbf{ASSET}, \emph{\textbf{A}daptive \textbf{S}peech \textbf{SE}gmen\textbf{T}ation}, to learn a boundary policy with a trainable pooling function for adaptive encoder feature compression.

Our contributions are:
\textbf{1)} ASSET, a framework that combines a learned segmenter with a residual pooler to adaptively produce compact audio representations while retaining task-oriented information;
\textbf{2)} a joint optimization paradigm that uses RL to learn segment boundaries from task feedback, while training the pooler and decoder through direct teacher-forcing;
and \textbf{3)} ASR experiments showing that the system works across languages, while exploratory non-ASR studies revealing how learned policies differ across tasks and speaking style.

\section{Related Work}
\label{sec:related_work}

Decoder-only speech systems shorten continuous acoustic prompts with CTC collapse \cite{wu2023decoder,zhou2025cjst} or fixed pooling \cite{bhati2025compression}. EFIN improves training efficiency by pretraining the encoder before LLM integration \cite{mu2025efin}. AdaTS adaptively merges tokens using a pairwise-similarity threshold \cite{sannigrahi2026adats}. By contrast, ASSET learns the boundary policy from task performance feedback while retaining every frame in a monotonic partition.

Adaptive reduction predates LLM-based ASR. Continuous integrate-and-fire learns soft monotonic alignments \cite{dong2020cif}, while extreme fixed reduction compresses transducer encoder outputs \cite{prabhavalkar2024extreme}. Jiang et al.\ jointly optimize phone error and frame rate with RL in hybrid ASR \cite{jiang2021vfr}; REBORN rewards boundaries using predicted-phoneme perplexity in unsupervised ASR \cite{tseng2024reborn}. In contrast, our approach optimizes boundaries against its LLM decoder's transcription loss and pools each segment with an order-aware residual in a supervised manner.

\section{Method}
\label{sec:method}

The goal of ASSET is to compress the frame sequence from a speech encoder before passing it to an LLM decoder (Fig.~\ref{fig:asset_overview}).
We extract contextual speech features with a frozen encoder and use a learned boundary policy to partition them into contiguous, variable-duration segments. We then summarize each segment with a residual recurrent pooler and prepend the resulting audio tokens to the decoder input to condition text generation.

\subsection{Adaptive segmentation}
\label{sec:ar}

Let $H=(h_1,\ldots,h_T)$ denote the encoder features. Our policy model outputs a logit $a_t=f_\theta(H,b_{<t})$ and boundary probability $p_t=\sigma(a_t)$. It then samples the boundary indicator $b_t\mid H,b_{<t}\sim\operatorname{Bernoulli}(p_t)$. $b_t=1$ starts a segment at frame $t$, while $b_t=0$ continues the current segment. With the first segment starting at frame one:
\begin{equation}
 s_t=1+\sum_{j=2}^{t} b_j, \qquad
 M=1+\sum_{t=2}^{T} b_t ,
 \label{eq:segments}
\end{equation}
where $s_t$ is the index of the segment that assigns frames to $M$ contiguous segments, as illustrated in Figure~\ref{fig:asset_overview}.
We denote the frame kept ratio as $\rho=M/T$ and the token rate as $f_{\mathrm{audio}}=\rho f_{\mathrm{enc}}$, for encoder rate $f_{\mathrm{enc}}$.

The policy distribution is
\begin{equation}
 \pi_\theta(b\mid H)
 =\prod_{t=2}^{T}p_\theta(b_t\mid H,b_{<t}).
 \label{eq:policy}
\end{equation}
In our implementation, the policy model is a CNN model or a transformer with local attention. Each conditional decision follows a Bernoulli distribution. However, the hidden states do condition on previous decisions $b_{<t}$, which makes the sampling process causal rather than independent.
We will show in Section~\ref{sec:asr_results} how conditioning on previous boundary decisions affects the task performance.

\subsection{Residual segment pooling}

Contextual features generated by the encoder contain ordering information, but their mean is permutation-invariant within a segment. To preserve the ordering information more explicitly, we additionally apply a bidirectional gated recurrent unit (BiGRU) \cite{schuster1997bidirectional} to extract the information from a segment. Let $\mathcal{S}_m=\{t:s_t=m\}$ denote the set of frame indices assigned to segment $m$. The BiGRU's terminal forward and backward states $\overrightarrow{g}_m$ and $\overleftarrow{g}_m$ provide an order-explicit information to the segment mean:
\begin{equation}
 \overline{h}_m=\frac{1}{|\mathcal{S}_m|}\sum_{t\in\mathcal{S}_m}h_t, \qquad
 z_m=\overline{h}_m+W_r[\overrightarrow{g}_m;\overleftarrow{g}_m]+c_r .
 \label{eq:pooler}
\end{equation}
Initializing $W_r,c_r=0$ is equivalent to mean pooling exactly, so introducing the recurrent branch does not inject noise into the encoded audio representations. The mean retains every frame's encoded representation directly, while the continuation loss optimizes the learned residual. As with other speech language models, we apply a projection layer that maps the pooled representations $Z=(z_1,\ldots,z_M)$ into the LLM embedding space.

\subsection{Task-guided policy optimization}

Our goal is to train a boundary policy model that preserves task performance while shortening the audio prefix. For each utterance, it samples $K$ complete boundary sequences $b^{(k)}\sim\pi_\theta$. We then pool the encoder features accordingly and score the resulting prefixes with the current LLM decoder. Let $\ell_k$ denote the normalized teacher-forced negative log-likelihood of the target continuation. For ASR, this continuation is the reference transcript.

However, continuation loss alone provides no explicit incentive to compress the audio prefix. The policy could choose no downsampling to avoid information loss from compression. We therefore introduce a soft regularizer on the retained ratio $\rho_k=M_k/T$. Its upper bound penalizes lazy over-segmentations, while its lower bound penalizes excessive compression, which might lead to training collapses. We combine this regularizer with the continuation loss to define the reward:
\begin{align}
 \mathcal{P}_{\mathrm{band}}(\rho)
 &= [\rho_{\min}-\rho]_+^2+[\rho-\rho_{\max}]_+^2, \\
 R_k &= -\ell_k-\lambda\mathcal{P}_{\mathrm{band}}(\rho_k).
 \label{eq:return}
\end{align}
Here $[x]_+=\max(x,0)$ and $\lambda$ controls the regularization strength.

We use GRPO \cite{shao2024deepseekmath} to optimize the policy model. Specifically, we standardize their returns to obtain relative advantages:
\begin{equation}
 \widehat{A}_k=\frac{R_k-\overline{R}}{s_R+\epsilon},
 \qquad
 \overline{R}=\frac{1}{K}\sum_{k=1}^{K}R_k.
 \label{eq:advantage}
\end{equation}
Here $s_R$ is the standard deviation of the group returns and $\epsilon$ stabilizes normalization. Segmentations with above-average returns receive positive advantages, encouraging the policy to assign them higher probability. We implement on-policy optimization with one update per step:
\begin{equation}
 \mathcal{L}_{\mathrm{GRPO}}
 =-\frac{1}{K}\sum_{k=1}^{K}
 \frac{\widehat{A}_k}{T-1}
 \sum_{t=2}^{T}
 \log p_\theta(b_t^{(k)}\mid H,b_{<t}^{(k)}).
 \label{eq:grpo}
\end{equation}
We omit reference-policy KL regularization ($\beta=0$), as we aim to encourage the model to deviate from its starting point that has a higher frequency. On the same minibatch, we use cross-entropy $\mathcal{L}_{\mathrm{CE}}$ to train the pooler, projection layer, and decoder adapters. The two losses update disjoint parameter sets: GRPO trains the boundary policy through its decision probabilities, while cross-entropy trains other components given the segmentation.

\section{Experiments}
\label{sec:experiments}

\begin{table*}[t]
\caption{Test results (mean $\pm$ sample SD, three seeds). Bold: best WER per panel; lower ASSET rate in (a). $^\dagger$~uses test transcripts. (b) matches initialization and seven epochs. GRPO uses a token-rate band of 7.5--12.5~Hz.}
\label{tab:results}
\centering
\small
\setlength{\tabcolsep}{4.2pt}
\renewcommand{\arraystretch}{0.85}
\begin{tabular}{@{}llrrrr@{}}
\toprule
System & Configuration & \multicolumn{2}{c}{Audio tokens/s $\downarrow$} & \multicolumn{2}{c}{WER (\%) $\downarrow$} \\
 & & Clean & Other & Clean & Other \\ \midrule
\multicolumn{6}{@{}l}{\textit{(a) Full LibriSpeech-960h system comparison}} \\ \addlinespace[1pt]
No downsampling & --- & 50.0 & 50.0 & 4.34 $\pm$ 0.47 & 7.08 $\pm$ 0.19 \\
Fixed grid & $k=5$ & 10.0 & 10.0 & 4.37 $\pm$ 0.10 & 7.72 $\pm$ 0.18 \\
Phone-CTC & audio-only & 10.9 & 10.5 & 4.05 $\pm$ 0.08 & 7.61 $\pm$ 0.06 \\
Character oracle$^\dagger$ & char-CTC & 14.6 & 14.3 & 2.87 $\pm$ 0.07 & 5.66 $\pm$ 0.15 \\
ASSET (CNN) & first-order AR & 11.9 & 11.6 & \textbf{2.75 $\pm$ 0.05} & 5.67 $\pm$ 0.10 \\
ASSET (Transformer) & local-64 AR & \textbf{10.9} & \textbf{10.6} & 2.79 $\pm$ 0.02 & \textbf{5.63 $\pm$ 0.12} \\
\midrule
\multicolumn{6}{@{}l}{\textit{(b) Matched LibriSpeech-100h: pooling and policy learning}} \\ \addlinespace[1pt]
Fixed grid & $k=5$, fixed / mean & 10.0 & 10.0 & 5.42 $\pm$ 0.23 & 9.71 $\pm$ 0.30 \\
Fixed grid & $k=5$, fixed / BiGRU & 10.0 & 10.0 & 5.17 $\pm$ 0.05 & 9.70 $\pm$ 0.12 \\
Frozen policy & local-64 AR, fixed / BiGRU & 14.4 & 13.9 & 5.09 $\pm$ 0.34 & 9.52 $\pm$ 0.27 \\
ASSET (joint) & local-64 AR, GRPO / BiGRU & 11.2 & 10.8 & \textbf{4.95 $\pm$ 0.02} & \textbf{9.11 $\pm$ 0.18} \\
\bottomrule
\end{tabular}
\end{table*}

\subsection{Experimental setup}

\textbf{Data and models.} Our design studies use LibriSpeech \cite{panayotov2015librispeech} train-clean-100; main systems use all 960h (train-clean-100/360 and train-other-500). Frozen WavLM-Large \cite{chen2022wavlm} produces features at approximately 50 Hz and feeds Llama-3.2-1B-Instruct \cite{meta2024llama32} with rank-16 LoRA \cite{hu2022lora}. The Transformer policy model has four 256-D layers and 64-frame local boundary history; the CNN policy model uses first-order boundary history through a learned bias term using two 256-channel convolutional layers followed by an output layer. The one-layer BiGRU has 128 hidden units per direction.

\noindent\textbf{Training.} We use $K=4$ rollouts for policy sampling, AdamW \cite{loshchilov2019adamw}, and BF16 on one A100 per run. Learning rates are $5\times10^{-5}$ for the policy and pooler, and $2\times10^{-4}$ for the projection and LoRA adapters. We report three ASR seeds with greedy decoding.

\noindent\textbf{Baselines.} The no-downsampling baseline keeps all encoder outputs; fixed grids place a boundary every $k$ frames. The character-boundary oracle uses CTC alignments for downsampling with transcripts given during both evaluation and training. The Phone-CTC baseline generates segment boundaries using phoneme predictions (without the transcript) from a frozen CTC module trained for ASR.

\noindent\textbf{Optimization curriculum.} We first initialize the segmenter with a binary cross-entropy training phase using character-level CTC boundaries as targets. This cold start is analogous to supervised fine-tuning (SFT) in LLM post-training, where models are pretrained before being further tuned with RL~\cite{shao2024deepseekmath,xiao2026escapelanguagepriormitigating}. We then use cross-entropy loss to adapt the pooler and decoder with the segmenter frozen, warm-starting the system to work with the compressed features with dynamic rates. This avoids noisy reward signals in the early RL stage and therefore stabilizes the training. In the last stage, we update the segmenter with the GRPO loss while continuing to fine-tune the decoder with cross-entropy.

\begin{figure}[!t]
  \centering
  \includegraphics[width=\columnwidth]{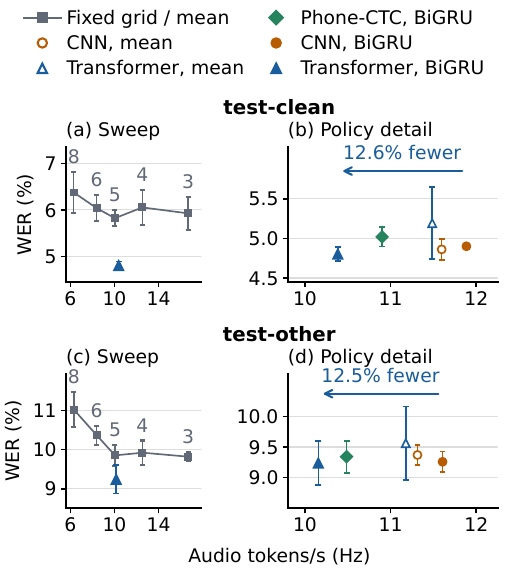}
  \caption{
  100h WER. Left: fixed strides $k$ and Transformer/BiGRU. Right: policy/pooler detail.
  }
  \label{fig:results_100h}
\end{figure}

\begin{table}[t]
\caption{Non-ASR test scores (\%, mean $\pm$ sample SD, three seeds). Rate bands: emotion 2--5, speaker count 3--6~Hz.}
\label{tab:nonasr_results}
\centering
\small
\setlength{\tabcolsep}{3pt}
\renewcommand{\arraystretch}{0.85}
\begin{tabular}{@{}lrrrr@{}}
\toprule
 & \multicolumn{2}{c}{Emotion} & \multicolumn{2}{c}{Speaker count} \\
\cmidrule(lr){2-3}\cmidrule(l){4-5}
Segmentation & Macro-F1 & Hz & Acc. & Hz \\
\midrule
No downsampling & $40.4 \pm 6.0$ & 50.0 & $72.4 \pm 6.9$ & 50.0 \\
Fixed grid ($k=2$) & $42.2 \pm 2.3$ & 25.0 & $77.3 \pm 5.9$ & 25.0 \\
Fixed grid ($k=5$) & $43.4 \pm 3.3$ & 10.0 & $73.5 \pm 2.1$ & 10.0 \\
Fixed grid ($k=10$) & $41.9 \pm 3.3$ & 5.0 & $73.6 \pm 3.4$ & 5.0 \\
\midrule
Frozen ASR policy & $43.1 \pm 0.4$ & 8.6 & $72.9 \pm 1.1$ & 10.6 \\
ASSET (Transformer) & $43.8 \pm 4.4$ & 4.1 & $72.4 \pm 1.2$ & 9.8 \\
\bottomrule
\end{tabular}

\end{table}

\begin{table}[t]
\caption{AISHELL-1 test CER and measured rates. $^\dagger$~uses test transcripts. Rate band: 3.25--10.0~Hz.}
\label{tab:mandarin_results}
\centering
\small
\setlength{\tabcolsep}{4pt}
\renewcommand{\arraystretch}{0.9}
\begin{tabular}{@{}lrr@{}}
\toprule
Segmentation & Tokens/s & CER (\%) $\downarrow$ \\
\midrule
Fixed grid ($k=5$) & 10.1 & 9.17 $\pm$ 0.20 \\
Phone oracle$^\dagger$ & 7.1 & 12.73 $\pm$ 0.36 \\
Character oracle$^\dagger$ & 3.2 & 8.63 $\pm$ 0.07 \\
\midrule
ASSET (Transformer) & 7.7 & \textbf{7.53 $\pm$ 0.07} \\
\bottomrule
\end{tabular}

\end{table}

\subsection{ASR results}
\label{sec:asr_results}

\noindent\textbf{Recognition at 960h.} Table~\ref{tab:results}(a) compares systems trained on the full LibriSpeech-960h set. The transformer-based system achieves 2.79/5.63\% WER at 10.9/10.6 Hz, about 4.6$\times$ prefix reduction and \textbf{8--9\% fewer tokens than the CNN system} at comparable WER. Both approach the character oracle with fewer tokens. The transformer system improves on phone-CTC by 1.25/1.98 WER points at similar rates.

\noindent\textbf{Ablation at 100h.} The broader sweep varies fixed downsampling rates and the policy models (Fig.~\ref{fig:results_100h}) introduced in Section~\ref{sec:ar}.
Our transformer system reaches 4.80/9.24\% WER at 10.4/10.2 Hz, about \textbf{13\% fewer tokens} than the CNN system.

\noindent\textbf{Pooling and policy ablations.} Table~\ref{tab:results}(b) holds decoder initialization and the training budget fixed. The first two rows use identical $k=5$ boundaries and change only the pooling method: BiGRU improves test-clean WER by 0.25 points, with test-other WER nearly unchanged. The last two rows start from the same transformer policy and train the BiGRU pooler and decoder in both cases; the segmenter stays frozen in one arm and receives GRPO updates in the other. We observe that joint GRPO training improves on the frozen policy by 0.14/0.41 WER while using about 3 fewer tokens/s. Full-path RTF measurements on the 100h systems show 2.8×/2.7× higher batched throughput for CNN/Transformer ASSET than no downsampling at each system’s largest stable tested batch (32 versus 16), demonstrating that decoder savings outweigh segmentation and pooling overhead.

\noindent\textbf{Mandarin ASR.} We train separate systems on AISHELL-1's 151h training split~\cite{bu2017aishell}, using frozen Chinese HuBERT-Large~\cite{hsu2021hubert,tencentchinesehubert} with the same Llama decoder. Baselines use fixed $k=5$, character-CTC, and Montreal Forced Aligner phone alignments~\cite{mcauliffe2017mfa}. We use phone-boundary initialization, followed by 24,000 updates with $\rho\in[0.065,0.20]$.
Unlike English, Mandarin characters typically represent syllables~\cite{zhou2018syllable}. Phone alignment therefore produces a higher token rate than character alignment (\MandarinPhoneRate{} versus \MandarinCharacterRate{}, Table~\ref{tab:mandarin_results}). ASSET reaches \MandarinCER\% CER at \MandarinRate{} tokens/s, outperforming both oracles with token rate comparable to phones.

\subsection{Exploratory non-ASR extensions}
\label{sec:nonasr}

We adapt the same cold-start initialization separately for emotion recognition (CREMA-D emotion \cite{cao2014cremad}) and speaker counting (synthetic LibriSpeech speaker counting data), using task-specific rate bands (Table~\ref{tab:nonasr_results}). We see that for emotion recognition, the system roughly halves the frozen policy's token rate with a similar mean macro-F1.

We study how policies learned for different tasks segment the same speech, and how a fixed policy responds to different expressions of the same text. On the Expresso \cite{nguyen2023expresso} data, we compare different policies using MFA phones with a 20 ms tolerance. We observe that different policies prefer different transitions for the same speech: emotion and speaker counting policies both keep about $3/4$ of ASR boundaries entering vowels from consonants, but retain \EmotionVCRetention\% and \CountVCRetention\%, respectively, when leaving vowels for consonants. We also compare the emotion segmenter on recordings of the same text spoken by the same speaker with different expressions. The boundary agreement with the default rendition is 67.7--75.2\% F1 across happy, sad, and confused speech, indicating substantial overlap with some variation in the policies.
These observations suggest that the policy adjusts token allocation based on how an utterance is spoken.

\section{Conclusion}
\label{sec:conclusion}

We propose ASSET, a framework for compressing speech LLM inputs through learned segmentation and residual recurrent pooling. We train a boundary policy with GRPO using task feedback, and the pooler and decoder with teacher-forced cross-entropy. Our approach achieves WER comparable to a transcript-assisted character-boundary oracle with fewer audio tokens on LibriSpeech-960h, while experiments on linguistically distinct Mandarin show similar benefits from our method. Exploratory non-ASR experiments reveal that segmentation varies across tasks and, within a single policy, with how the same utterance is expressed.

\newpage

 \section{Acknowledgement}
This work was supported by JSALT 2026 at Johns Hopkins University with funds from NSF CCRI Grant No. 2120435, Google DeepMind, JHU HLTCOE, JHU AI2AI and ACL.

We used AI tools to assist in generating some code snippets for the experiments reported in Section~\ref{sec:experiments}. The authors take full responsibility for the code, analyses, and reported results.

\section{Compliance with Ethical Standards}
This study used only existing speech datasets; our institution determined that no additional ethics approval was required.

\bibliographystyle{IEEEbib-linked-compact}
\bibliography{references}

\end{document}